\documentclass[lettersize,journal]{IEEEtran}
\usepackage{amsmath,amsfonts}
\usepackage{algorithmic}
\usepackage{algorithm}
\usepackage{array}
\usepackage[caption=false,font=normalsize,labelfont=sf,textfont=sf]{subfig}
\usepackage{textcomp}
\usepackage{stfloats}
\usepackage{url}
\usepackage{verbatim}
\usepackage{graphicx}
\usepackage{cite}

\usepackage{amssymb}
\usepackage[whole]{bxcjkjatype}
\usepackage{bm}
\usepackage{enumerate}
\usepackage{subfig}
\usepackage{xspace}
\usepackage{amssymb}
\newcommand{\bhline}[1]{\noalign{\hrule height #1}}  
  
\newcommand{\ctext}[1]{\raise0.2ex\hbox{\textcircled{\scriptsize{#1}}}}
\allowdisplaybreaks

\usepackage{color}

\newcommand{\OursIntg}{AOSA\xspace}

\newcommand{\approxOursIntg}{AOSA-approx\xspace}

\begin{document}

\title{Adaptive occlusion sensitivity analysis \\ for visually explaining video recognition networks}

\author{Tomoki Uchiyama, Naoya Sogi, Satoshi Iizuka, Koichiro Niinuma, and Kazuhiro Fukui
\thanks{T. Uchiyama, S. Iizuka and K. Fukui are with the Department of Computer Science, University of Tsukuba, Tsukuba, Japan. Email: uchiyama@cvlab.cs.tsukuba.ac.jp; iizuka@cs.tsukuba.ac.jp; kfukui@cs.tsukuba.ac.jp}
\thanks{N. Sogi was with the Department of Computer Science, University of Tsukuba. He is now with the Visual Intelligence Research Laboratories, NEC corporation, Japan. Email: naoya-sogi@nec.com} 
\thanks{K. Niinuma is with Fujitsu Research of America, PA, USA. Email: kniinuma@fujitsu.com}
}




\maketitle

\begin{abstract}
This paper proposes a method for visually explaining the decision-making process of video recognition networks with a temporal extension of occlusion sensitivity analysis, called Adaptive Occlusion Sensitivity Analysis (AOSA).
The key idea here is to occlude a specific volume of data by a 3D mask in an input 3D temporal-spatial data space and then measure the change degree in the output score. The occluded volume data that produces a larger change degree is regarded as a more critical element for classification.
However, while the occlusion sensitivity analysis is commonly used to analyze single image classification, applying this idea to video classification is not so straightforward as a simple fixed cuboid cannot deal with complicated motions.
To solve this issue, we adaptively set the shape of a 3D occlusion mask while referring to motions.
Our flexible mask adaptation is performed by considering the temporal continuity and spatial co-occurrence of the optical flows extracted from the input video data. 
We further propose a novel method to reduce the computational cost of the proposed method with the first-order approximation of the output score with respect to an input video.
We demonstrate the effectiveness of our method through various and extensive comparisons with the conventional methods in terms of the deletion/insertion metric and the pointing metric on the UCF101 dataset and the Kinetics-400 and 700 datasets.
\end{abstract}


\section{Introduction}
\label{intro}

\begin{figure*}[t]
    \centering
    \includegraphics[width=\linewidth]{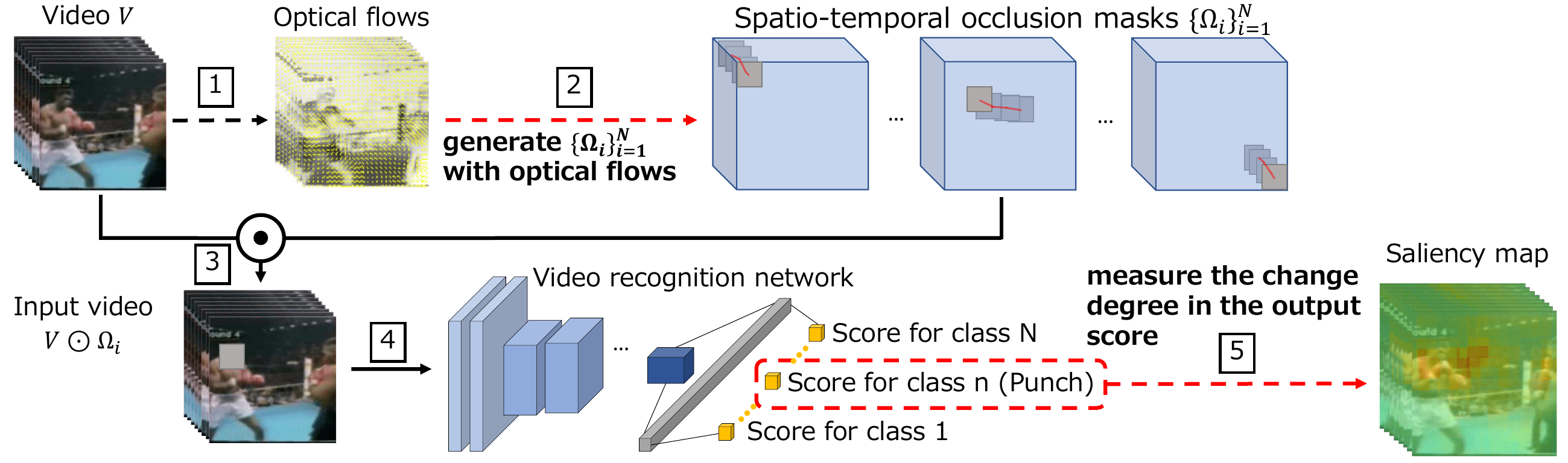}
    \caption[The framework of Adaptive~Occlusion~Sensitivity~Analysis~(AOSA)]{The framework of Adaptive~Occlusion~Sensitivity~Analysis~(AOSA). Our approach tracks objects using optical flow so that a spatio-temporal occlusion mask can occlude the same object throughout the video. A saliency map is generated by measuring the change degree in the output score when the occluded video is input to the video recognition network.}
    \label{fig:proposed}
\end{figure*}

\IEEEPARstart{T}{his} paper proposes an adaptive occlusion sensitivity analysis for visualizing and understanding the decision-making process of video recognition networks. Our occlusion sensitivity analysis provides informative sensitivity maps that indicate which parts of an input video are more important for explaining network predictions.

With increased attention to the high ability of deep neural networks (DNN), how to explain the DNN predictions has become a fundamental problem in both theoretical study and practical applications, particularly in serious tasks directly related to human life such as medical diagnosis and accident investigation of an autonomous vehicle~\cite{Tjoa2021}. Many types of methods have been proposed to provide such an explanation, mainly for convolutional neural networks (CNN) taking a single image as input~\cite{gradcam, osm, rise, ep}, as will be surveyed in the next section. 

This paper focuses on a method using the occlusion sensitivity analysis (OSA)~\cite{osm}, which has been widely used for visually explaining CNN predictions due to its simple idea and high expandability. Our basic idea motivated by the OSA is to occlude a specific volume of data by a 3D mask in an input 3D temporal-spatial data space and then measure the change degree in the output score as shown in Fig.\ref{fig:proposed}.
We regard an occluded volume that produces a more significant change in the score as more critical data for the predictions. 

OSA is simple and easy to implement. However, it is not straightforward to be incorporated into the video recognition network architecture since it has been originally developed to explain the prediction process of the standard 2D CNN. 
For example, one may come up with a way of extending a rectangle mask along the temporal direction to a cuboid mask in an input 3D spatio-temporal space as a simple extension of the OSA. However, this naive approach does not work as expected when a target object moves, as a fixed cuboid mask cannot continue to occlude the moving object throughout the video.

To address this issue, we propose to change the shape of a 3D occlusion mask adaptively to the complicated motions of a target object, considering the temporal continuity and spatial co-occurrence of the motions. To this end, we first extract the optical flows \cite{opticalflow,pyflow_original,pyflow} from an input video and then move each occlusion mask set to an attention region detected in the first frame of the video. In this way, the movement trace of an occlusion mask, which we call a spatio-temporal occlusion mask $\Omega$, forms a blended tube mask with a rectangular shape, as shown in the upper-right of Fig. \ref{fig:proposed}. 
This 3D mask is the base of our method. 

We need to consider multiple spatio-temporal occlusion masks to simultaneously handle multiple objects with more complicated motions. For this purpose, we group multiple occlusion masks with similar positions and motions into the same class and then integrate a set of masks into one mask $\hat\Omega$ with a larger volume size. In this way, we generate a more effective 3D mask. We call this new analysis $\it Adaptive~Occlusion~Sensitivity~Analysis~(AOSA)$.

Furthermore, we discuss two valid options for further enhancing our analysis method. First, we describe applying the conditional sampling method~\cite{conditional} to fill more statistically valid intensity values in each occlusion mask instead of using a constant value.
Next, we describe the computational cost of our method.
Conducting inference many times over a whole video causes an additional problem of high computational cost, although its algorithm is simple. 
We introduce an approximation computation of the change score to reduce the cost. Considering a network as a certain function that transforms an input video to a class score, we approximate the function in the first order using its Taylor expansion, where the first-order partial derivative of the function with respect to an input video can be obtained by the automatic differentiation through back-propagation. 

In the experiment section, we first conduct a qualitative evaluation of our sensitive map in comparison with that of various conventional methods. Next, we compare the effectiveness of our method with the conventional methods in terms of the deletion/insertion metric and
the pointing metric~\cite{rise}, which have been widely used as valid indexes, on the video classification on the UCF101 dataset~\cite{ucf101} and Kinetics-400 and 700 datasets~\cite{kinetics400,kinetics700}.

Our main contributions are summarized as follows.
\begin{itemize}
\item[(1)] We propose an adaptive occlusion sensitivity analysis (AOSA) for explaining video recognition networks.
\item[(2)] We introduce an approximation computation for calculating the change degree in the class score to reduce the high computational cost.
\item[(3)] We demonstrate the effectiveness of our method through an extensive comparison with the conventional methods in terms of the deletion/insertion metric and the pointing metric on UCF101 and Kinetics-400 and 700 datasets.
\end{itemize}

The paper is organized as follows. 
In Section II, we describe the related methods. 
In Section III, we discuss the proposed method. First, we explain the basic idea of our adaptive occlusion sensitivity analysis. Then, we construct a method for explaining video recognition networks. Further, we introduce an approximate computation of occlusion sensitivity analysis using Taylor expansion.
In Section IV, we demonstrate the effectiveness of our method through evaluation experiments. Finally, Section V concludes the paper. 

This paper is an extended and enhanced version of our earlier publication \cite{uchiyama2023visually}. Although some ideas in this paper have appeared in the publication, the present paper contains a more practical formulation of our framework by introducing a speed-up technique based on the first-order Taylor approximation of the occlusion sensitivity analysis. Furthermore, we provide more comprehensive evaluation experiments, including performance comparison on additional datasets (Kinetics-400 and Kinetics-700) using other models (SlowFast and TimeSFormer) and ablation studies.

\section{Related Work}

\subsection{Explanation of 2D CNN predictions}
Most interpretability methods provide the information for explaining the CNN predictions, that is, the decision-making process by a saliency map indicating which elements/regions of an input image contribute to the output score. 
For 2D-CNN taking a single image as an input, many types of methods for generating such a saliency map have been proposed \cite{partialgrad,revbp,shap,pmlr-v70-koh17a,feraud2002methodology,arrieta2020explainable,baehrens2010explain,lime,survey21}. These are categorized into three types of methods: perturbation-based, activation-based and gradient-based.

As a simple yet effective perturbation-based method, we review a method using the occlusion sensitivity analysis (OSA). The idea of the OSA-based method is straightforward. First, it measures the slight variation of the class score to occlusion in different regions of an input image using small perturbations of the image. Then, the resultant score variation of each region is summarized as a saliency map called a sensitivity map of the input image. In the sensitivity map, the local image regions with significant variation are emphasized as the part that positively contributes to the class score. Accordingly, the occlusion sensitivity map can provide helpful information for understanding what image features contribute to a final decision and further implies why the network fails the classification.
Meaningful Perturbation~\cite{meaningful} and Extremal Perturbation~\cite{ep} generate a saliency map opposite the OSA. 
These methods occlude the elements/regions with fewer contributions while remaining the elements/regions with significant contributions.

Grad-CAM~\cite{gradcam} has been well known as one of the popular activation-based methods. The Grad-CAM generates a saliency as the weighted sum of the convolutional feature maps, where the gradient is used as the weight of each feature map. 

Although the Grad-CAM has been widely used for explaining CNN predictions, it has a limitation in spatial resolution since the spatial resolution of the Grad-CAM is determined by the low resolution of the last layer. For example, the resolution of Grad-CAM is only 7-by-7 pixels when using the GooglNet. Thus, the resolution of a saliency map from Grad-CAM is usually much lower than an occlusion map. 

For the gradient-based methods, Deep LIFT~\cite{deeplift} and Guided Backprop~\cite{guidedbp} have been proposed for explaining CNN predictions. These methods obtain the contribution of each input element by applying back-propagation. They can provide a pixel-wise fine saliency map. However, it is often difficult to visually understand the meaning of the map.

\subsection{Extensions for video recognition networks}
There are a few methods for explaining the decision-making process of networks taking videos as input. 
A naive approach is applying methods such as OSA and Grad-CAM to video recognition networks without any modification just by replacing 2D-matrix with 3D-tensor data as an input. For example, an extension of Grad-CAM, Grad-CAM++~\cite{gradcam++}, has been applied to 3D-CNN for explaining the process of action recognition. 
However, such naive approaches cannot work well, as will be demonstrated in the experimental section, as they do not have a mechanism for explicitly handling the temporal relationship in video data. 

Several methods considering the temporal relationship have been proposed.
Saliency tube~\cite{saliencytubes} generates a spatio-temporal saliency map as an extension of CAM~\cite{cam}, where the feature map of the last layer is weighted with the coefficients from the prediction layer.
It has been shown that the  separation of the information from Grad-CAM into spatial and temporal information works effectively for a more detailed explanation~\cite{Hiley2020}.
SWAG-V~\cite{swag-v} enhances the framework SWAG~\cite{swag} by averaging and smoothing a saliency map at the super-pixel level.

Like our approach, Saptio-Temporal Extremal Perturbation (STEP)~\cite{step} is also a perturbation-based method. This method tackles how to handle the temporal relationship effectively, unlike other methods, which simply apply existing techniques designed for 2D-CNN. STEP extends the framework of Extremal Perturbation ~\cite{ep} to handle the task of explaining predictions of video recognition networks. The key idea here is to solve an optimization problem with the smoothness constraint on successive temporal saliency maps. Although STEP can use the temporal information well, its computational cost is high due to the heavy optimization. Besides, STEP tends to be affected by some noise to output an unclear map.

\section{Proposed Method}
In this section, we extend the idea of OSA to a visual explanation of video recognition networks. In a 3D spatio-temporal data composed by an input video, the temporal direction is simultaneously occluded in addition to the spatial direction.

\subsection{Generating masks with optical flows}
\label{sec:single}

\begin{figure}[t]
    \centering
    \includegraphics[width=0.9\linewidth]{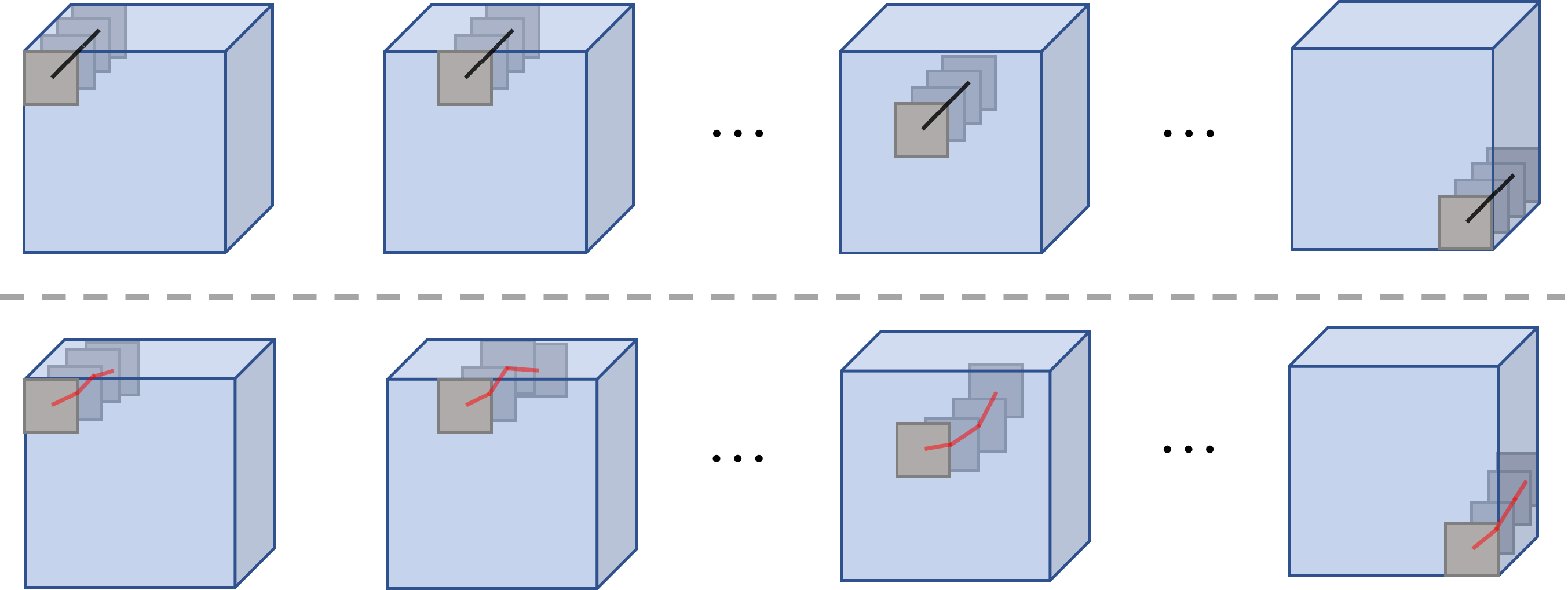}
    \caption[A simple extension of OSA for 3D and the proposed method]{A simple extension of OSA for 3D (top) and the proposed method (bottom). The simple 3D OSA occludes a cuboid. On the other hand, the proposed method occludes a curved tubular rectangular shape (spatio-temporal occlusion mask) along the displacement vectors of the optical flows.}
    \label{fig:3d_occ}
\end{figure}

As mentioned in Section \ref{intro}, a simple extension of OSA is to occlude a part of the 3D spatio-temporal data with a fixed cuboid, as shown in the top of Fig. \ref{fig:3d_occ}. Unfortunately, this method occludes a different object in each frame when the objects are moving. Therefore, it is difficult to produce a meaningful visualization map. 
To overcome this limitation, the proposed method generates occlusion masks while following the motion of the target object based on the optical flow~\cite{opticalflow,pyflow_original,pyflow}, as shown in the bottom of Fig. \ref{fig:3d_occ}. 

Let an input color video with $T$ frames be $V \in \mathbb{R}^{T\times H \times W \times 3}$ and $I_t \in \mathbb{R}^{H \times W \times 3}$ be the $t$-th frame image.
Occlusion masks $\{M_i^t\in \{0, 1 \}^{H \times W \times 3}\}$ are generated by the following process:
\begin{enumerate}[(1)]
\item $N$ anchor points $\{p^{(1)}_i\in\mathbb{R}^2\}_{i=1}^N$ for $I_1$ are equally spaced vertically and horizontally every $s$ pixels across the entire image.
\item A rectangle mask $M_i^1 \in \{0, 1 \}^{H \times W \times 3}$ is set up to occlude a $h \times w$ region centered at each anchor point $p_i^{(1)}$. The masked region of $M_i^t$ is set to 0 and the other regions are set to 1.
\item In the image of the second frame $I_2$, each anchor point $p^{(2)}_1, \dots, p^{(2)}_N$ is moved by the optical flow.
\item If an anchor point moves out of the input image, its tracking stops at that frame, and no further occlusion is conducted.
\item A rectangle mask $M_i^2 \in \{0, 1 \}^{H \times W \times 3}$ is set up to occlude a $h \times w$ region centered at each anchor point $p_i^{(2)}$. 
\end{enumerate}

By repeating the above process until the final frame $T$, we obtain $T\times N$ rectangle masks $\{M_i^t\}$.
Each spatio-temporal occlusion mask $\Omega_{i}\in\{0, 1\}^{T\times H\times W\times 3}$ is generated by arranging rectangle masks $\{M_i^t\}_t$ over the temporal domain, resulting in that $N$ masks $\{\Omega_i\}_{i=1}^N$ being obtained.
A spatio-temporal occlusion mask represents the moving of an anchor point or an object. 

When the output score changes significantly by applying a spatio-temporal occlusion mask $\Omega_{i}$, the corresponding occluded regions are visualized as important for a decision of a network.

\subsection{Consideration of co-occurrence}

\begin{figure}[t]
    \centering
    \includegraphics[width=0.75\linewidth]{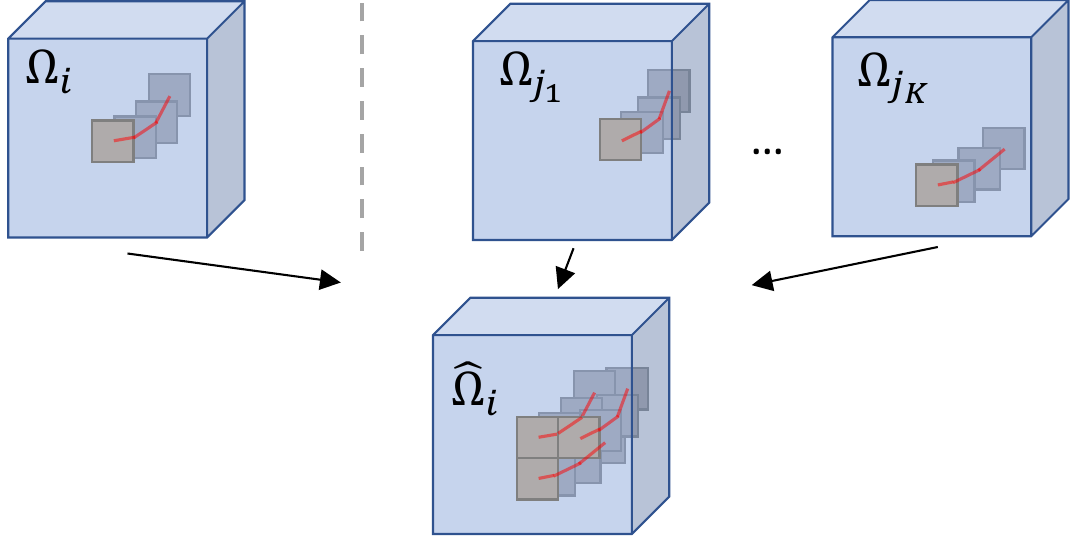}
    \caption[Generation of integration mask]{Generation of integration mask. The highly co-occurring masks with a mask $\Omega_i$ are selected by Eq~(\ref{eq:co}). Then, the integrated mask $\hat{\Omega}_i$ is calculated by the element-wise product among $\Omega_i$ and $\{\Omega_{j_k}\}_{{j_k}\in{\cal K}_i}$.
    }
    \label{fig:composite_masks}
\end{figure}

In the previous Section \ref{sec:single}, $N$ spatio-temporal occlusion masks $\{\Omega_{i}\}$ are applied independently, and the change degrees in the output scores are measured without considering their relationship. However, spatio-temporal occlusion masks capturing the same object's motion have co-occurrence, which is essential information for the classification.
Therefore, the influence of highly co-occurred motions on the classification should be investigated.
To this end, we measure the degree of co-occurrence and then integrate masks with highly co-occurred motions into a single integrated mask $\hat{\Omega}$, as shown in Fig.\ref{fig:composite_masks}.

\vskip\baselineskip
\noindent $\blacksquare$ Co-occurrence between occlusion masks: 
There are many ways to measure degrees of co-occurrence between spatio-temporal occlusion masks. Since the proposed method uses optical flow for generating masks, considering motion, we measure co-occurrence based on the variation pattern of displacement vectors $\{u_{i}\}$ obtained by optical flow, i.e., movement of anchor points $\{p_i^{(t)}\in\mathbb{R}^2\}_t$. 

We define the co-occurrence between $i'$ and $i$th spatio-temporal occlusion masks $\Omega_{i}, \Omega_{i'}$ as follows:
\begin{equation}
    Co(\Omega_{i}, \Omega_{i'}) = {u_{i} \cdot u_{i'}} / ({\|u_{i} \| \|u_{i'}\|}),
    \label{eq:co}
\end{equation}
where $u_{i} =  [(p_{i}^{(2)}-p_{i}^{(1)})^\top, \dots , (p_{i}^{(T)}-p_{i}^{(T-1)})^\top]^\top \in \mathbb{R}^{2(T-1)}$ is a displacement vector.

\vskip\baselineskip
\noindent $\blacksquare$ Mask integration: 
We integrate spatio-temporal occlusion masks by the co-occurrence $Co$. First, we obtain the top $K$ masks $\{\Omega_j\}_{j\in {\cal K}_i}$ that co-occur with the $i$th mask $\Omega_{i}$. We then apply the element-wise product among $\Omega_i$ and $\{\Omega_j\}_{j\in {\cal K}_i}$ to generate the integration mask $\hat{\Omega}_i$, such that all target regions by the masks are occluded. The integration masks $\{\hat{\Omega}_i\}_{i=1}^N$ are used in place of the original masks $\{\Omega_{i}\}_{i=1}^N$.

Finally, we apply occlusion by calculating the element-wise product to the integrated mask $\hat{\Omega}_i$ and the input video $V$, i.e. $\hat{\Omega}_i \odot V$. Then, we input the occluded video into a classification model and measure the output score of the target class. We estimate that the occluded regions are important for the classification if the output score becomes low.
With reference to ~\cite{rise}, the visualization map $S$ is given as the weighted sum of the mask $\hat{\Omega}_i$, where the weights are defined with the output score $f(\hat{\Omega}_i \odot V)$, as follows:
\begin{equation}
    S=\frac{1}{N} \sum^N_{i=1} f(\hat{\Omega}_i \odot V) \cdot {\hat{\Omega}}_i,
\end{equation}
where $f$ is a classification model.

\begin{figure}[t]
    \centering
    \includegraphics[width=0.85\linewidth]{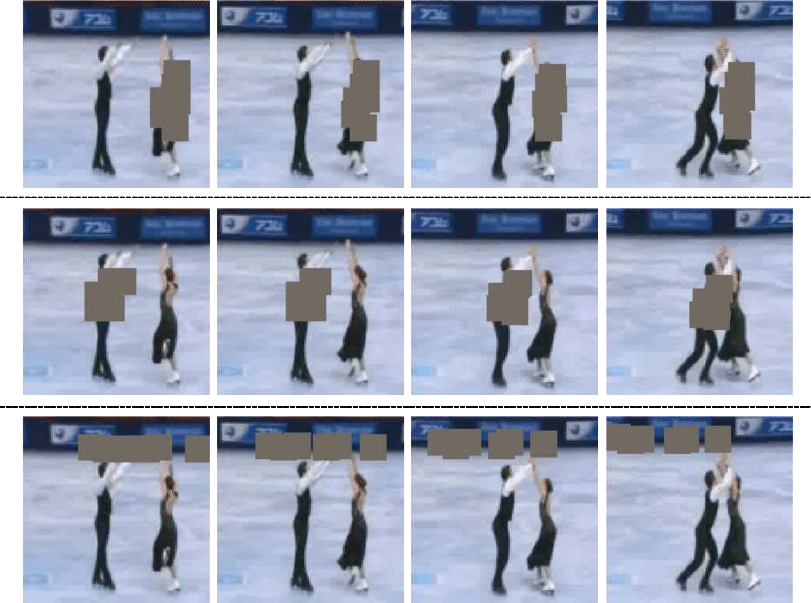}
    \caption[A video with integration masks]{
    A video with integration masks ($K=5$).
    Each row shows the occluded frames by an integrated mask.
    }
    \label{fig:mask_examples}
\end{figure}

Fig. \ref{fig:mask_examples} shows examples of a video with integration masks. We can see that the same objects like people can be occluded, by using the co-occurrence.

\vskip\baselineskip
In the following, we refer to this method as Adaptive Occlusion Sensitivity Analysis (AOSA).

\subsection{Conditional sampling}
Conditional sampling~\cite{conditional} provides a more stable analysis by using a statistically sound method to rigorously measure the prediction changes when certain elements of the input are removed. The prediction changes may make no sense in some cases if the elements are naively replaced with a uniform value. Instead of filling up occluded regions with such a uniform value, conditional sampling introduced the calculation method of more appropriate replaced values $v$ of occluded regions to OSA. The replaced value is randomly sampled from the normal distribution, whose parameters are estimated from patches around the occluded region.

In the conditional sampling, it first extracts patches $\{\hat{\bm{\mathrm{x}}}_i^j\}$ from around an anchor point $p_i$, where each patch includes the pixel at the $p_i$, and its size is same as the occluded regions.
Then, the mean $\mu$ and variance $\Sigma$ are calculated from the patches without original occlusion patch.
A replaced value $v$ is randomly sampled from ${\cal N}(\mu, \Sigma)$.
A visualization map is obtained by using an average output score with multiple randomly sampled values.
An average score can be written as $\sum_{v_i}q(v_i|\mu,\Sigma)f(x')$, where $q(v_i|\mu,\Sigma)$ is the probability density function, $x'$ is the occluded input with a replaced value $v_i \in \mathbb{R}^{h \times w}$, and $f$ is a neural network.

\subsection{Speedup by approximation}
\label{approx}
The disadvantage of the occlusion sensitivity analysis is the high computational cost since many occluded inputs are required to feed a neural network. Besides, applying the conditional sampling increases the computational cost dramatically as the number of inferences can increase proportionately to the number of random sampling. 
To address this issue, we assume that the change to the input caused by occlusion is small, as an occluded region is typically small.
This assumption enables us to approximate the inference of the neural network by a first-order Taylor expansion. As a result, this approximation realizes fast computation of the output values by using a simple linear calculation.
As shown later, this approximation makes the computational cost of the conditional sampling almost the same as that of OSA.

\subsubsection{First-order approximation of OSA}
Let $f:\mathcal{X} \to \mathbb{R}$ be a function from the input space $\mathcal{X}$ to the output score $\mathbb{R}$, which is modeled by a neural network. In the following, we assume that the input is a vector for simplicity. When a small value $\delta_x$ is added to the input $x$, the first-order approximation by Taylor expansion of $f(x+\delta_x)$ in the neighborhood of $x$ is as follows:
\begin{equation}
\label{Taylor}
f(x+\delta_x) \simeq f(x) + \left(\frac{\partial f}{\partial x}|_x \right)^\top (x + \delta_x -x).
\end{equation}
From equation (\ref{Taylor}), we can calculate the approximate inference of a neural network by using only $f(x)$ and the partial derivative $\frac{\partial f}{\partial x}|_x$ of $f$ with respect to $x$. $\frac{\partial f}{\partial x}|_x$ is obtained by back-propagating~\cite{back-prop} the output score to the input $x$ after $f(x)$ is obtained through the neural network. In such way, the approximation of $f(x+\delta_x)$ can be computed with a single forward $f(x)$ and the corresponding backward pass. The back-propagation can be easily computed using the automatic differentiation with deep learning libraries such as PyTorch~\cite{pytorch}.

Next, we explain how to compute the approximate inference of the occluded input by the neural network using equation (\ref{Taylor}). Let $m$ be the mask, and $v$ be a matrix including replaced values. The occluded input $g(x;m,v)$ can be represented as $g(x;m,v)=x \odot m + (1-m) \odot v$. Therefore, the inference of the occluded input by the neural network can be represented as $f(g(x;m,v))$.
From equation (\ref{Taylor}), the first-order approximation of $f(g(x;m,v))$ is as follows:
\begin{align}
f(g(x;m,v)) \simeq f(x) &+ \left(\frac{\partial f}{\partial x}|_x \right)^\top (g(x;m,v)-x) \notag\\
&={\hat{f}}(g(x;m,v)).
\label{eq:firstapprox}
\end{align}
Finally, the importance $S_{m}$ of the occluded region by $m$ is the difference from the output score of the original input as follows:
\begin{equation}
\label{importance}
    S_{m} = f(x) - {\hat{f}}(g(x;m,v)).
\end{equation}

\subsubsection{Approximation using conditional sampling}
\label{conditional}

According to~\cite{conditional}, equation (\ref{importance}) for the conditional sampling is as follows:
\begin{equation}
\label{conditional_importance}
    S_{m} = f(x) - \sum_{v_i}q(v_i)f(g(x;m,v_i)).
\end{equation}
Here, since $g(x;m,v_i)-x=(1-m) \odot (v_i-x)$, the first-order approximation of equation (\ref{conditional_importance}) is as follows:
\begin{align}
\label{eq:approx_conditional}
    & f(x) - \Sigma_{v_i}q(v_i)f(g(x;m,v_i)) \notag\\
    & \simeq f(x) - \left\{ f(x) + \sum_{v_i}q(v_i)J_x^\top (g(x;m,v_i)-x) \right\} \notag\\
    & = -J_x^\top \sum_{v_i}\left\{q(v_i) (1-m) \odot (v_i-x) \right\} \notag\\
    & = -J_x^\top (1-m)\odot\{\sum_{v_i}{q(v_i) v_i} - x\sum_{v_i}q(v_i)\} \notag\\
    & = -J_x^\top(1-m) \odot (\mu-x), 
\end{align}
where $J_x$ is the partial derivative of $f(x)$ with respect to $x$, $\sum_{v_i}q(v_i)v_i$ is an expectation $\mu$ of the probability distribution, and $\sum_{v_i}q(v_i)=1$.
Therefore, the number of the neural networks' inferences does not relate to the number of random sampling. 
This makes the computational cost of the conditional sampling almost the same as the OSA.

\subsubsection{Use of a gradient technique}
\label{relu} 
Gradient based methods often produce signed values. It is difficult to determine whether positive or negative values are more important~\cite{smoothgrad}. Therefore, selecting only one of the positive or negative values~\cite{shrikumar2016not}, or taking absolute values~\cite{simonyan2013deep, smoothgrad}, are done to make the visualization more understandable. 

In sensitivity analysis, we are interested only in whether occlusion lowers the output. To extract only negative elements, we modify the approximation in the equation (\ref{eq:approx_conditional}) using Rectified Linear Unit (ReLU) as follows:
\begin{equation}
S_m = \sum_i -ReLU( J_{x_i}(1-m_i) \odot (\mu-x_i)).
\end{equation}
The introduction of ReLU significantly improves evaluation metrics that focus on elements with high values in the saliency map.

We refer to the proposed method, which introduces the above approximation and stabilization techniques, as AOSA-approx.

\section{Evaluation Experiments}
In this section, we evaluate the proposed methods through comparison with the conventional explanation methods. To this end, we conduct qualitative evaluation by direct visualization of saliency maps and quantitative evaluation using deletion and insertion metrics~\cite{rise} and S-PT~\cite{step, excitation}.

\subsection{Experiments settings}
We use the three action recognition datasets, 1) UCF101~\cite{ucf101}, 2) Kinetics-700, and 3) Kinetics-400, for evaluation. The details of each dataset and the experimental settings are as follows.

\vskip\baselineskip
\noindent $\blacksquare$ UCF101:
The UCF101 dataset has a total of 3783 test videos in 101 categories. We use ResNet50-based R3D~\cite{3dresnet} and R(2+1)D~\cite{r2p1d} as classification models. Saliency maps are generated after fine-tuning the networks ~\cite{3dresnet_3}, pre-trained on the Kinetics-700~\cite{kinetics700}. The input videos are clipped and resized to 16~frames $\times$ 112 $\times$112~pixels. For AOSA, we place the anchor points equally spaced every $s=8$ pixels in the first frame. The total number of anchor points is, in this case, $N=196(=(112/8)^2)$. Then, a rectangle occlusion mask of $16\times16$ pixels is placed at each anchor point. 

\noindent $\blacksquare$ Kinetics-700:
The Kinetics-700 dataset has 700 categories. We select ten videos in each category from the validation data for evaluation. We use ResNet50-based R3D~\cite{3dresnet} and R(2+1)D~\cite{r2p1d} as classification models. The input videos are clipped and resized to 16~frames $\times$ 112 $\times$112~pixels. For AOSA, we place the anchor points equally spaced every $s=4$ pixels in the first frame. Then, a rectangle occlusion of $8\times8$ pixels is placed around each anchor point. 

\noindent $\blacksquare$ Kinetics-400:
The Kinetics-400 dataset has 400 categories. We select ten videos in each category from the validation data for evaluation. We use SlowFast~\cite{slowfast} and TimeSFormer~\cite{timesformer} as classification models~\cite{mmaction2}. The input videos are clipped and resized to 32~frames $\times$ 256 $\times$256~pixels for SlowFast and 8~frames $\times$ 224 $\times$224~pixels for TimeSFormer. For AOSA, we place the anchor points equally spaced every $s=10$ pixels for SlowFast and every $s=8$ pixels for TimeSFomer in the first frame. Then, a rectangle occlusion mask of $20\times 20$ pixels for SlowFast and $16\times 16$ pixels for TimeSFormer is placed at each anchor point.

\vskip\baselineskip

We generate an integration mask $\hat{\Omega}_i$ by integrating the target mask $\Omega_i$ and three highly co-occured masks.
For AOSA-approx, patches are extracted from the region that is enlarged by 4 pixels around the occlusion mask.

We compare the proposed method with Grad-CAM~\cite{gradcam}, occlusion sensitivity analysis (OSA)~\cite{osm}, SmoothGrad~\cite{smoothgrad}, and STEP~\cite{step}. Grad-CAM and OSA are applied by extending the 2D matrix data of images to the 3D tensor data of videos.
In TimeSFomer, we also use Attention Rollout~\cite{rollout}, a visualization method for transformers, as a comparison method.
For OSA, we set the occlusion size to half the number of frames in the time direction and the size of 2\% of the entire image in the spatial direction. For STEP, the visualization method of the video recognition network, we use the default parameters~\cite{step}.

\subsection{Evaluation metrics}
We use two types of metrics from different viewpoints for the quantitative evaluation.

First, we use deletion and insertion metrics~\cite{rise}, which evaluate how faithfully the saliency map represents the inferences of the model. Deletion evaluates performance based on how quickly the model prediction probability reduces when pixels are deleted in the order of importance in the saliency map. Conversely, insertion evaluates performance based on how quickly the model prediction probability increases when pixels are inserted in order of importance. Specifically, both metrics use the area under the curve (AUC) where the horizontal axis is the percentage of pixels deleted or inserted, and the vertical axis is the output probability. We also provide the over-all score following~\cite{groupcam}, which can be calculated by \textit{AUC(insertion)−AUC(deletion)}, and evaluates both metrics comprehensively.
In this experiment, pixel deletion and insertion are performed in the same 28 iterations as in~\cite{swag-v}.

Second, we also use the spatial pointing game (S-PT) metric~\cite{step, excitation} that evaluates how well the explanation matches the human interpretation. For this evaluation, we use UCF101-24~\cite{ucf101}, which is a part of UCF101 annotated with bounding boxes indicating the area in which humans act. We perform this evaluation with the R3D model and the R(2+1)D model. In this experiment, following~\cite{step}, one hit is recorded when a 7-pixel radius circle centered at the maximum value of the saliency map intersects the bounding box in each frame. The hit rate in the entire dataset is defined as the evaluation score of S-PT. A good explanation will have a high score value assuming that the model learns the humans acting as a feature.

\subsection{Qualitative results}

\begin{figure*}[t]
 \centering
  \includegraphics[width=\linewidth]{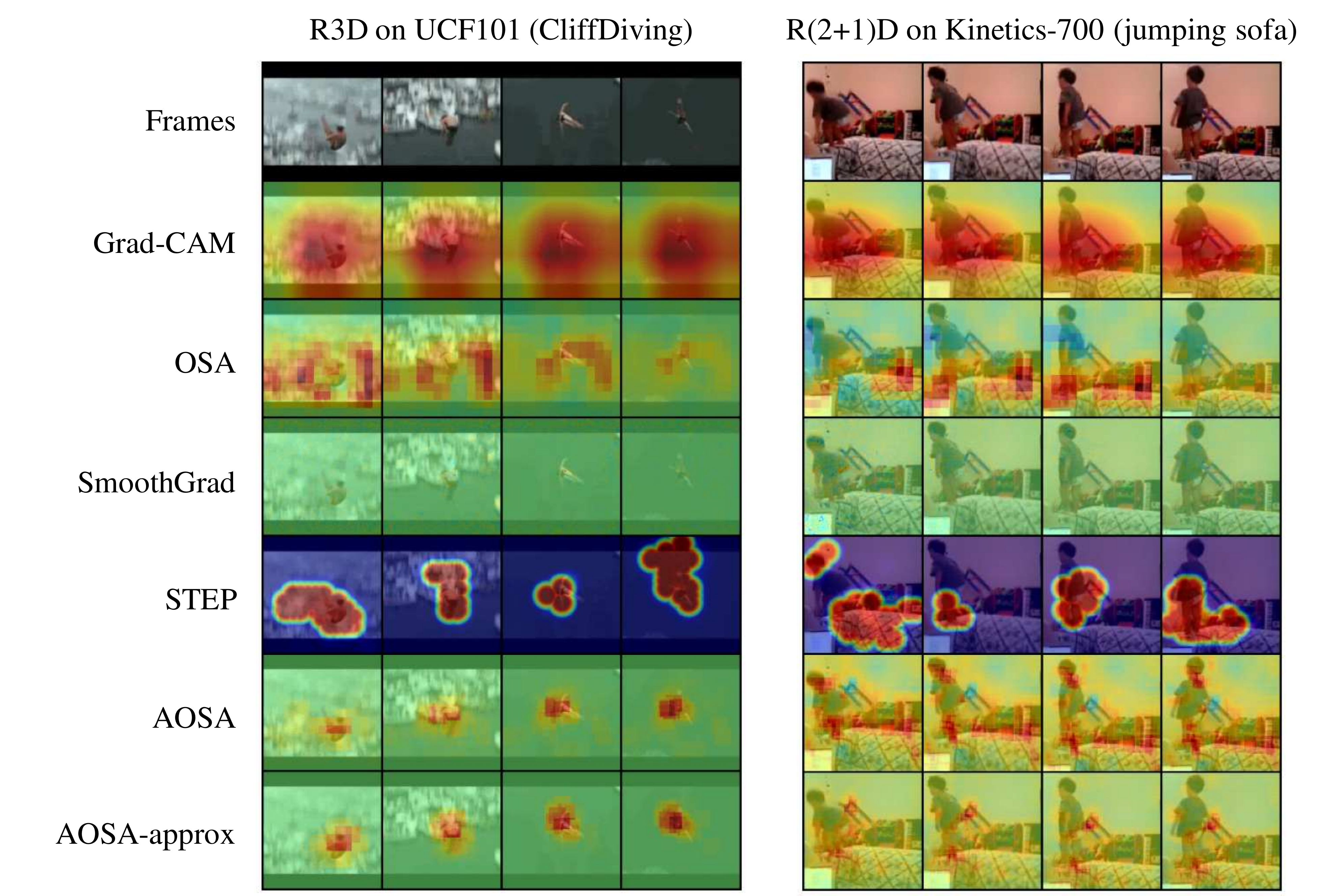}
  \caption[Visualization results of saliency maps in R3D and R(2+1)D]{Visualization results of saliency maps in R3D and R(2+1)D. The proposed methods generate stable and smooth saliency maps.}
  \label{fig:results1}
\end{figure*}
\begin{figure*}[t]
 \centering
  \includegraphics[width=\linewidth]{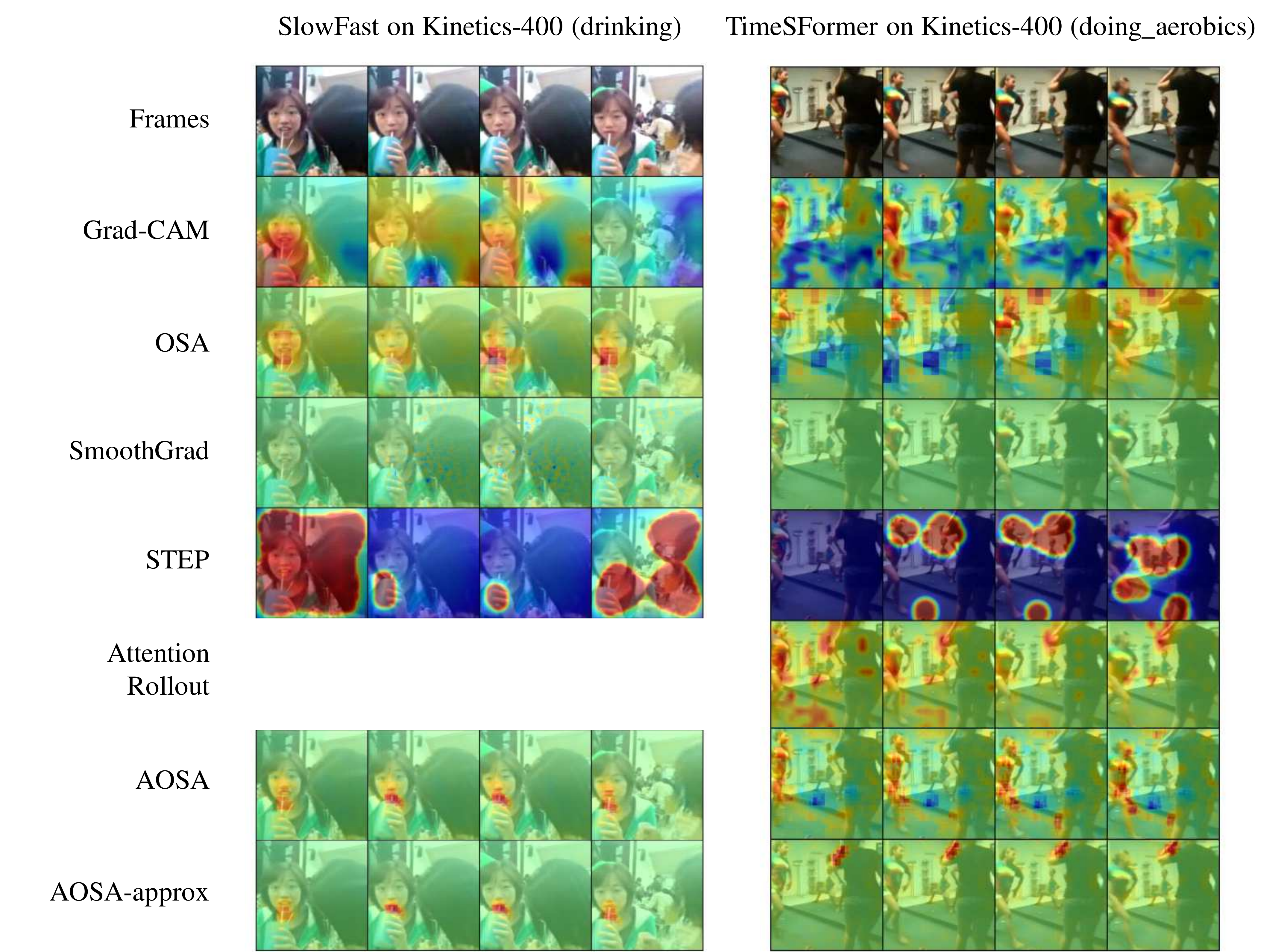}
  \caption[Visualization results of saliency maps in SlowFast and TimeSFormer]{Visualization results of saliency maps in SlowFast and TimeSFormer.}
  \label{fig:results2}
\end{figure*}

Figures \ref{fig:results1} and \ref{fig:results2} show the visualization results of the saliency maps using each model. We select four frames of the video. In Fig. \ref{fig:results1}, the left side shows the results of the CliffDiving class video in R3D, and the right side shows the results of the jumping sofa class video in R(2+1)D. In Fig. \ref{fig:results2}, the left side shows the results of the drinking class video in SlowFast, and the right side shows the results of the doing aerobics class video in TimeSFormer. 

These figures demonstrate Grad-CAM, OSA, and SmoothGrad fail to capture time-series variations. This is expected since these approaches do not consider the temporal structure of the video. Additionally, the maps generated by Grad-CAM on R3D and R(2+1)D are resized from a resolution of $1\times 4 \times 4$, which makes it difficult to precisely determine the network's focus points. Although STEP correctly captures several distinct regions, such as people and drinks, there appears to be inconsistency in the temporal direction.

Unlike these existing approaches, our proposed method accurately captures class-specific features such as the diving person and the jumping person, and the time-series variation of the maps is smooth.

\subsection{Quantitative results}

Tables \ref{table:del-ins_ucf}-\ref{table:del-ins_k400} show the evaluation results using the deletion and insertion metrics on UCF101, Kinetics-700, and Kinetics-400, respectively. According to ~\cite{swag-v}, the saliency maps that accurately capture important individual pixels receive high evaluation scores in terms of deletion metric. In contrast, the saliency maps that capture cohesive regions together receive high evaluation scores in terms of insertion metric. We confirm the same tendency in this experiment. 
In R3D and R(2+1)D, SmoothGrad performs well in the deletion. In contrast, Grad-CAM works well in the insertion. The difference may be explained by that SmoothGrad tends to produce limited local regions of high importance, while Grad-CAM provides comparatively large ones. The different characteristics may match well with each metric, respectively.
STEP's poor performance may be due to the fact that it generates saliency maps that are not well coherent in space and time. 

The proposed method shows competitive results in the deletion and insertion metrics compared to the conventional methods. These results show that the proposed method explains the process of making-decision of the network accurately and understandably. Regarding the over-all score, the proposed method has the best comprehensive performance. Furthermore, the proposed method retains its advantage over the conventional methods even when approximate calculations are introduced.

In Table~\ref{table:del-ins_k400}, AOSA-approx outperforms the base of AOSA, unlike in the other experiments. The characteristics of the target networks, SlowFast and TimeSFormer, may be able to explain this result.
We shall remember that AOSA-approx uses the partial derivative of an input image to calculate the importance. That suggests that AOSA-approx has a mechanism to enhance the edges of the image since the partial derivative corresponds to some kind of edge filter. 
This characteristic matches well the recently reported observation that SlowFast and TimeSFormer have similar ones. It is experimentally reported that SlowFast emphasizes not only static information but also motion information in an input video compared to other video recognition models \cite{ilic22appearancefree}. As suggested by the optical flow calculation, we need to enhance edges to stably extract the motion information from an input video, as the motion can be very unstable in the regions without clear edges. The transformer-based methods focus selectively on the information of edges and contours compared to CNN-based methods \cite{NEURIPS2021_c404a5ad}. These observations imply the reason why AOSA-approx works effectively together with these networks. In fact, we confirmed that AOSA is better than AOSA-approx when we use the R(2+1)D instead of them.

Then, we show the performances in terms of S-PT in Table \ref{table:S-PT}. The proposed methods achieve the best performance among the compared methods in S-PT. In particular, the proposed method significantly improves the performance of the original OSA. 

We can see that AOSA-approx is better than AOSA. This result may be related to the partial derivative function as the experimental result in Table \ref{table:del-ins_k400}. The metric of S-PT evaluates whether the importance of each mask could locate the critical spatial regions, which are defined by using manually annotated bounding boxes. In general, such bounding boxes can have many edges. That may help AOSA-approx improve further the performance of AOSA.

\begin{table}[t]
  \caption{Deletion and insertion scores on UCF101. For deletion, lower is better ($\downarrow$). For insertion, higher is better ($\uparrow$). For over-all score, higher is better ($\uparrow$). The best scores are in bold, while the second best scores are in underline.}
  \label{table:del-ins_ucf}
  \centering
  \begin{tabular}{l|cc|cc}
  \bhline{1.5pt}  
  \multicolumn{1}{c|}{} & \multicolumn{2}{c|}{Deletion ($\downarrow$) / Insertion ($\uparrow$)} & \multicolumn{2}{c}{Over-all ($\uparrow$)}\\
  Method & R3D & R(2+1)D & R3D & R(2+1)D\\
  \hline
  Grad-CAM~\cite{gradcam} & 0.206 / \underline{0.632} & 0.245 / \underline{0.674} & 0.426 & 0.429 \\
  OSA~\cite{osm} & 0.160 / 0.585 & 0.225 / 0.644 & 0.424 & 0.419 \\
  SmoothGrad~\cite{smoothgrad} & \textbf{0.058} / 0.357 & \textbf{0.056} / 0.275 & 0.299 & 0.219 \\
  STEP~\cite{step} & \underline{0.146} / 0.560 & \underline{0.150} / 0.607 & 0.414 & 0.457 \\

  \OursIntg & 0.158 / \textbf{0.646} & 0.182 / \textbf{0.688} & \textbf{0.489} & \textbf{0.507} \\
  \approxOursIntg & 0.193 / 0.621 & 0.198 / 0.658 & \underline{0.428} & \underline{0.460} \\
  \bhline{1.5pt}  
  \end{tabular}
\end{table}

\begin{table}[t]
  \caption{Deletion and insertion scores on Kinetics-700.}
  \label{table:del-ins_k700}
  \centering
  \begin{tabular}{l|cc|cc}
  \bhline{1.5pt}  
  \multicolumn{1}{c|}{} & \multicolumn{2}{c|}{Deletion ($\downarrow$) / Insertion ($\uparrow$)} & \multicolumn{2}{c}{Over-all ($\uparrow$)}\\
  Method & R3D & R(2+1)D & R3D & R(2+1)D\\
  \hline
  Grad-CAM~\cite{gradcam} & 0.088 / 0.305 & 0.119 / 0.353 & 0.217 & 0.235 \\
  OSA~\cite{osm} & 0.048 / \underline{0.310} & 0.076 / \textbf{0.378} & \underline{0.261} & \underline{0.301} \\
  SmoothGrad~\cite{smoothgrad} & \textbf{0.019} / 0.141 & \textbf{0.020} / 0.112 & 0.123 & 0.092 \\
  STEP~\cite{step} & 0.058 / 0.275 & 0.065 / 0.324 & 0.217 & 0.259 \\

  \OursIntg & \underline{0.045} / \textbf{0.326} & \underline{0.057} / \underline{0.369} & \textbf{0.282} & \textbf{ 0.312} \\
  \approxOursIntg & 0.066 / 0.301 & 0.073 / 0.340 & 0.234 & 0.267 \\
  \bhline{1.5pt}  
  \end{tabular}
\end{table}

\begin{table*}[t]
  \caption{Deletion and insertion scores on Kinetics-400.}
  \label{table:del-ins_k400}
  \centering
  \begin{tabular}{l|cc|cc}
  \bhline{1.5pt}  
  \multicolumn{1}{c|}{} & \multicolumn{2}{c|}{Deletion ($\downarrow$) / Insertion ($\uparrow$)} & \multicolumn{2}{c}{Over-all ($\uparrow$)}\\
  Method & SlowFast & TSFormer & SlowFast & TSFormer\\
  \hline
  Grad-CAM~\cite{gradcam} & 0.146 / 0.186 & 0.162 / 0.233 & 0.040 & 0.071 \\
  OSA~\cite{osm} & 0.167 / \underline{0.214} & 0.260 / \textbf{0.263} & 0.047 & 0.003 \\
  SmoothGrad~\cite{smoothgrad} & \textbf{0.041} / 0.101 & \textbf{0.109} / 0.184 & 0.060 & \textbf{0.075} \\
  STEP~\cite{step} & 0.237 / \textbf{0.256} & 0.237 / \underline{0.256} & 0.018 & 0.018 \\
  Attention Rollout~\cite{rollout} & - & 0.186 / 0.229 & - & 0.043 \\
  
  \OursIntg & 0.121 / 0.197 & 0.150 / 0.218 & \underline{0.076} & 0.069 \\
  \approxOursIntg & \underline{0.119} / 0.205 & \underline{0.141} / 0.213 & \textbf{0.086} & \underline{0.072} \\
  \bhline{1.5pt}  
  \end{tabular}
\end{table*}

\begin{table}[t]
  \caption{S-PT scores on UCF101-24 for R3D and R(2+1)D.}
  \label{table:S-PT}
  \centering
  \begin{tabular}{lcc}
  \bhline{1.5pt} 
  Method & R3D & R(2+1)D \\
  \hline
  Grad-CAM~\cite{gradcam} & 0.651 & 0.658 \\
  OSA~\cite{osm} & 0.522 & 0.597 \\
  SmoothGrad~\cite{smoothgrad} & 0.676 & 0.618 \\
  STEP~\cite{step} & 0.672 & 0.667 \\

  \OursIntg & \underline{0.682} & \underline{0.747} \\
  \approxOursIntg & \textbf{0.763} & \textbf{0.789} \\
  \bhline{1.5pt} 
  \end{tabular}
\end{table}

\subsection{Comparison of generation time}

In Table \ref{table:speed}, we show the mean computational time of different methods to generate a saliency map on R3D. 
In the proposed method, we measured the mean time with the intervals $s=8$ $(N=196)$ and $s=4$ $(N=784)$ of anchor points controlling the map resolution. Note that the resolution of the maps is equivalent to $14 \times 14$ for $s = 8$ and $28 \times 28$ for $s = 4$. 
Although the proposed method requires more processing time than the Grad-CAM, the proposed method is about 25 times faster than STEP with $s = 8$. This is because although both the proposed method and STEP need hundreds to thousands of inferences by the model, STEP requires the gradient computation of the network during optimization, unlike the proposed method.
Moreover, by approximating the inference of the network, the proposed method can further generate saliency maps about 2.4 times faster.
This is because the numbers of gradient computations and inferences reduce significantly thanks to the approximation compared with STEP.

\begin{table}[t]
  \caption{Mean computation time for generating a saliency map. These are the results measured in R3D.}
  \label{table:speed}
  \centering
  \begin{tabular}{lc}
  \bhline{1.5pt} 
  Method & Computation time (sec) \\
  \hline
  Grad-CAM~\cite{gradcam} & {\bf 0.021}\\
  OSA~\cite{osm} & 4.759\\
  SmoothGrad~\cite{smoothgrad} & 2.256  \\
  STEP~\cite{step} & 18.706\\
  \OursIntg (s=8) & 0.728\\
  \approxOursIntg (s=8) & \underline{0.302}\\
  \OursIntg (s=4) & 2.548\\
  \approxOursIntg (s=4) & 1.028\\ 
  \bhline{1.5pt} 
  \end{tabular}
\end{table}

\subsection{Ablation study}
\textbf{Integration masks.} 
To evaluate the impact of mask integration, we compare the performance of AOSA using a single mask versus integration masks. Table~\ref{table:ablation_integration} presents the results, assessed based on the overall score of deletion/insertion. The results confirm that integration masks can offer effective occlusion for the visualization.

\begin{table*}[t]
  \caption{Ablation study of Mask integration. We evaluate it with the over-all score.}
  \label{table:ablation_integration}
  \centering
  \begin{tabular}{l|cc|cc|cc}
  \bhline{1.5pt}  
  \multicolumn{1}{c|}{} & \multicolumn{2}{c|}{UCF101} & \multicolumn{2}{c|}{Kinetics-700} & \multicolumn{2}{c}{Kinetics-400}\\
  Method & R3D & R(2+1)D & R3D & R(2+1)D & SlowFast & TSFormer\\
  \hline
  AOSA with a single mask & 0.482 & 0.488 & 0.281 & 0.311 & 0.072 & 0.055\\
  AOSA with an integrated mask & \textbf{0.489} & \textbf{0.507} & \textbf{0.282} & \textbf{0.312} & \textbf{0.076} & \textbf{0.069}\\
  \bhline{1.5pt}  
  \end{tabular}
\end{table*}

\begin{figure}[t]
    \centering
    \includegraphics[width=\linewidth]{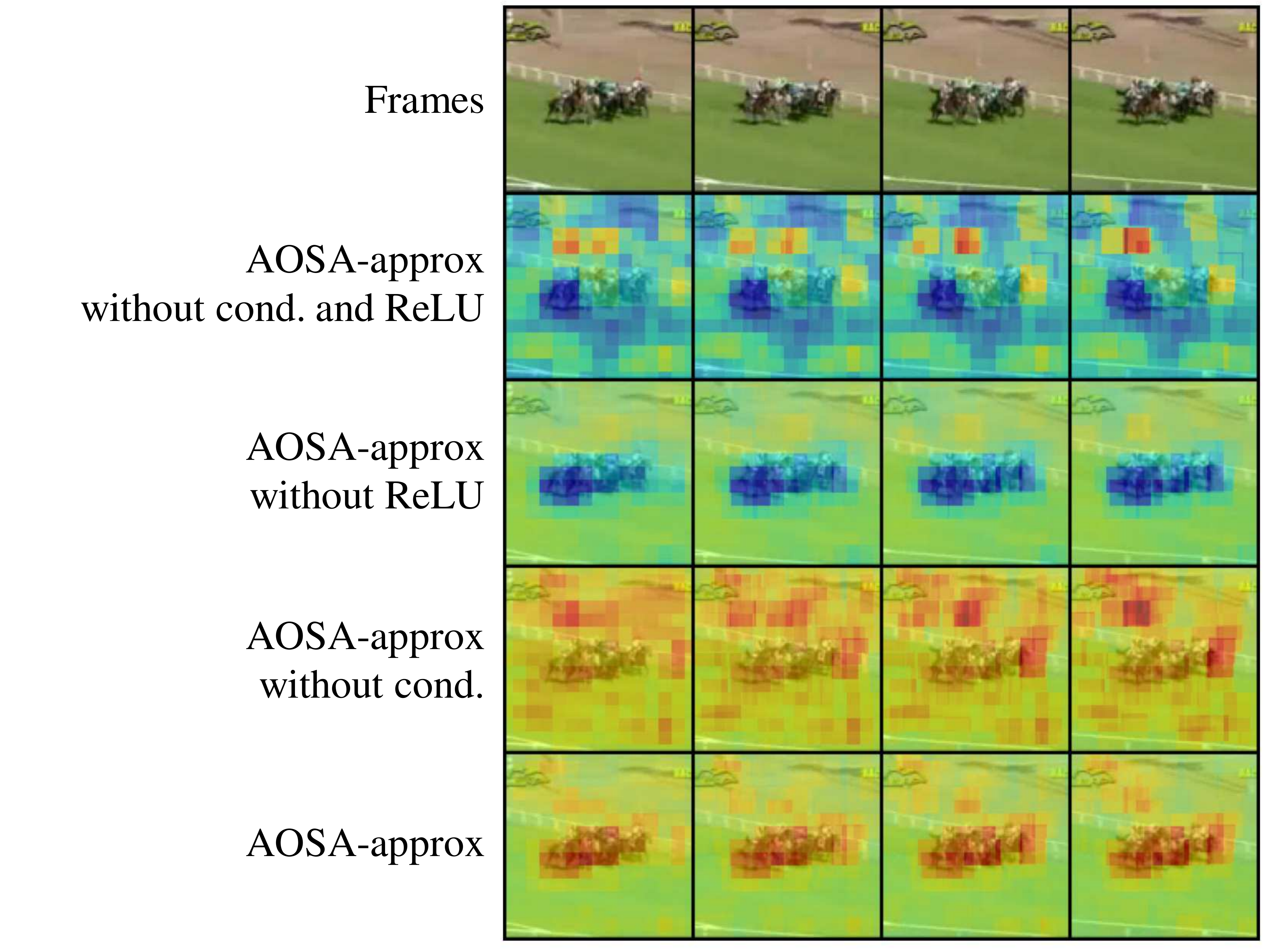}
    \caption{Saliency maps of \approxOursIntg for a HorseRace class video with and without conditional sampling and ReLU.}
    \label{fig:ablation_approx}
\end{figure}

\textbf{Stabilization for approximation.} Below, we confirm the effects of conditional sampling and ReLU introduced in Section \ref{conditional} and Section \ref{relu}, respectively. Fig. \ref{fig:ablation_approx} shows the saliency maps of \approxOursIntg with and without conditional sampling and ReLU.
Through conditional sampling, the responses to regions other than the horses are suppressed, which indicates increased stability in sensitivity analysis. Furthermore, we can confirm that ReLU passing only positive responses produces the consistent explanation.
In Table \ref{table:ablation_approx}, we investigate the quantitative results for conditional sampling and ReLU on UCF101. Applying them significantly improves each score. Based on these findings, we can conclude that the accuracy and understandability of the saliency maps are enhanced.

\begin{table}[t]
  \caption{Ablation study of \approxOursIntg. We evaluate it using R3D and R(2+1)D on UCF-101.}
  \label{table:ablation_approx}
  \centering
  \begin{tabular}{cc|cc|cc}
  \bhline{1.5pt} 
  \multicolumn{2}{c|}{Approach} & \multicolumn{2}{c|}{Deletion ($\downarrow$) / Insertion ($\uparrow$)} & \multicolumn{2}{c}{S-PT}\\
  cond. & ReLU & R3D & R(2+1)D & R3D & R(2+1)D\\
  \hline
   & & 0.244 / 0.548 & 0.277 / 0.564 & 0.489 & 0.485\\
   & \checkmark & 0.224 / 0.566 & 0.234 / 0.605 & 0.547 & 0.607\\
   \checkmark &  & 0.258 / 0.576 & 0.295 / 0.594 & 0.622 & 0.611\\
   \checkmark & \checkmark & \textbf{0.193} / \textbf{0.621} & \textbf{0.198} / \textbf{0.658}  & \textbf{0.763} & \textbf{0.789}\\
  \bhline{1.5pt} 
  \end{tabular}
\end{table}

\subsection{Sanity Check}
\begin{figure}[t]
    \centering
    \includegraphics[width=\linewidth]{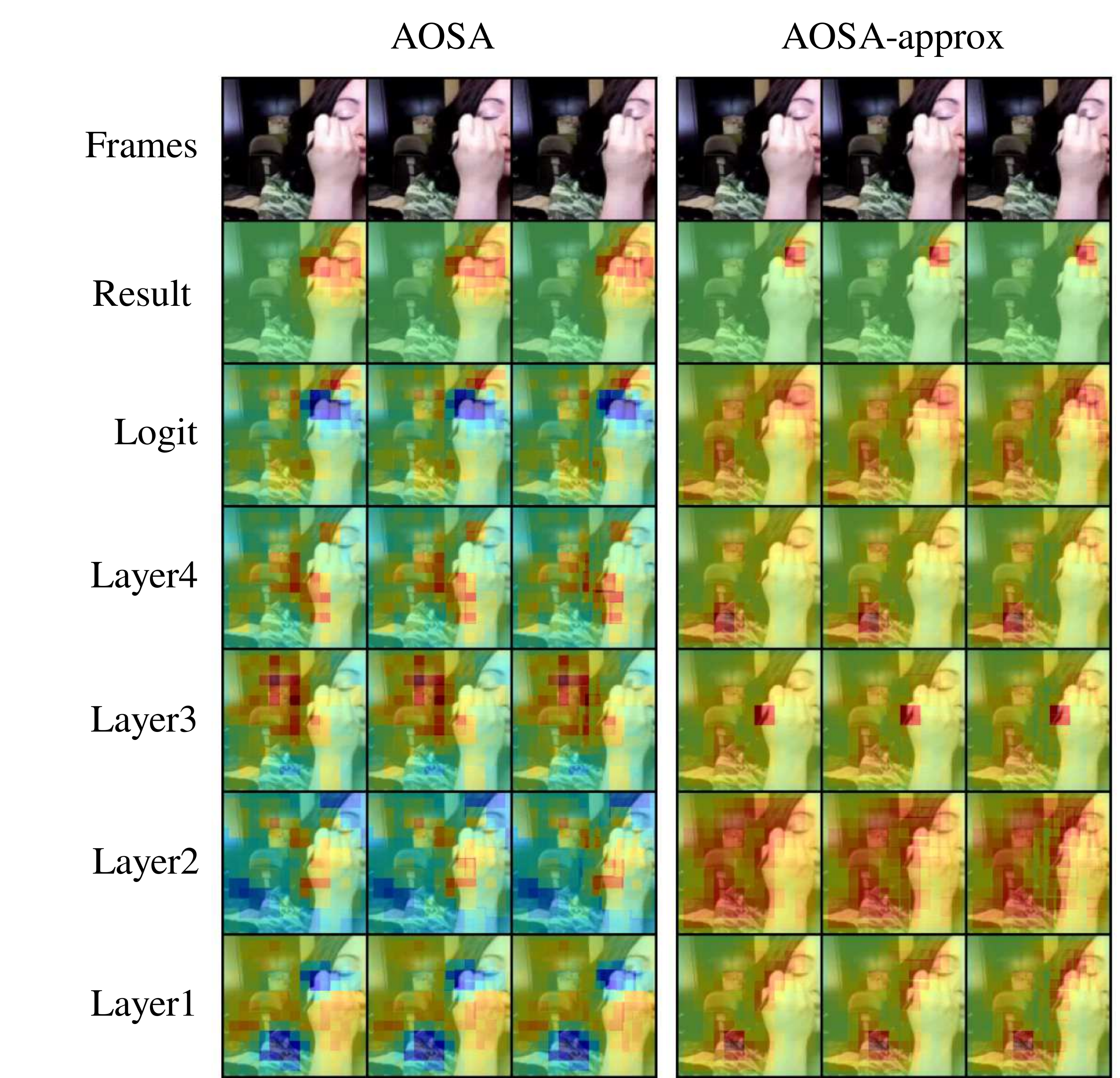}
    \caption{Sanity check results for AOSA and AOSA-approx on R3D.}
    \label{fig:sanity}
\end{figure}

We conduct sanity check~\cite{sanitycheck} to check the reliability of the explanations provided by AOSA and AOSA-approx for model parameters.
The sanity check is important to ensure that explanations are generated dependent on the model parameters.
Specifically, we apply cascade randomization, which progressively randomizes the parameters of each layer, to R3D from logit to layer 1. As shown in Fig. \ref{fig:sanity}, our experimental results indicate that the proposed methods are sensitive to model parameters, as the randomization of parameters causes the saliency maps to be destroyed.

\section{Conclusion}
In this paper, we have proposed an adaptive occlusion sensitivity analysis (AOSA) for visually explaining the decision-making process of video recognition networks. Our sensitivity analysis taking a video as input is a temporal extension of the original occlusion sensitivity analysis taking a single static image.
The novelty of our sensitivity analysis is to change the shape of the 3D occlusion mask adaptively to the complicated optical flows extracted from an input video. The co-occurrence of optical flows is used to generate occlusion masks with more effective shapes. Moreover, we have introduced a method for reducing the computational cost of the sensitivity analysis through its first-order approximation. The results of the evaluation experiment have demonstrated that the proposed method is quantitatively advantageous over the conventional methods and qualitatively provides clear explanations that are easy for users to understand.

\bibliographystyle{IEEEtran}
\bibliography{ref}

\newpage


\begin{IEEEbiography}[{\includegraphics[width=1in,height=1.25in,clip,keepaspectratio]{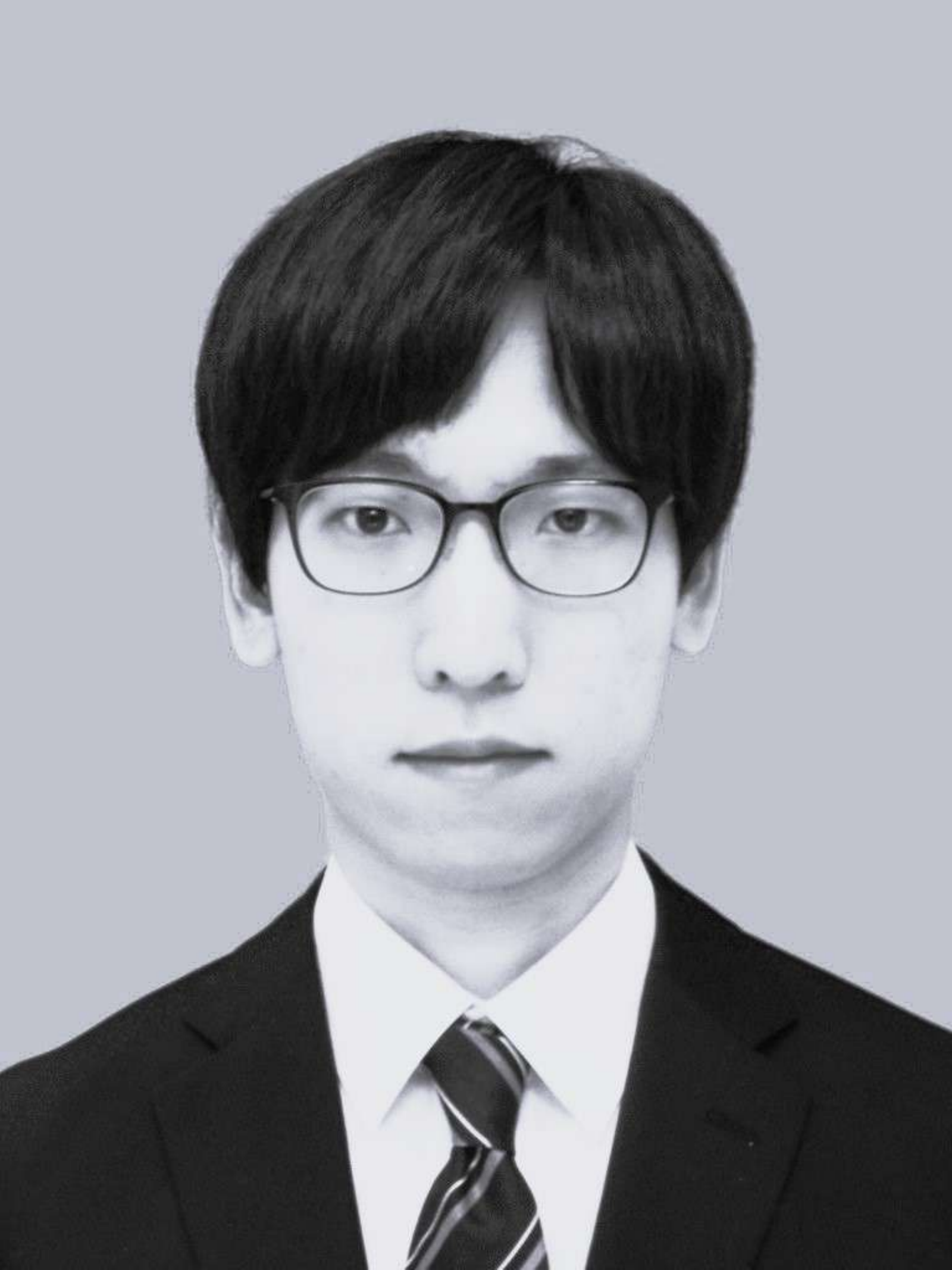}}]{Tomoki Uchiyama}
received Ms. Eng from the university of Tsukuba in 2023. His interests include
the theories of computer vision, pattern recognition and machine learning.
\end{IEEEbiography}
\vspace{11pt}

\begin{IEEEbiography}[{\includegraphics[width=1in,height=1.25in,clip,keepaspectratio]{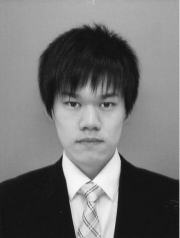}}]{Naoya Sogi}
received Ms. Eng. and Dr. Eng. from the University of Tsukuba in 2019 and 2022, respectively. He is currently a researcher at the NEC Visual Intelligence Research Laboratories. His interests include the theory of computer vision, pattern recognition, machine learning and applications of these theories.
\end{IEEEbiography}

\begin{IEEEbiography}[{\includegraphics[width=1in,height=1.25in,clip,keepaspectratio]{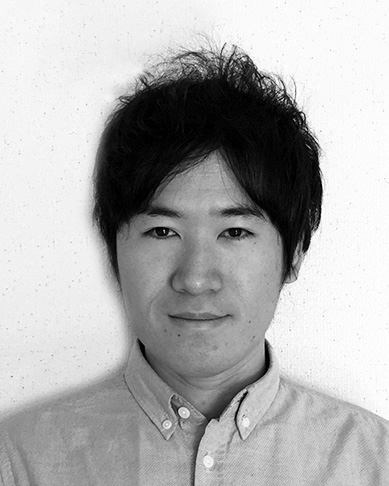}}]{Satoshi Iizuka}
is an associate professor at the University of Tsukuba's Faculty of Engineering, Information and Systems. He received his PhD in engineering from the University of Tsukuba. His research interests are in computer graphics and vision, including image processing and editing based on machine learning.
\end{IEEEbiography}
\vspace{11pt}

\begin{IEEEbiography}[{\includegraphics[width=1in,height=1.25in,clip,keepaspectratio]{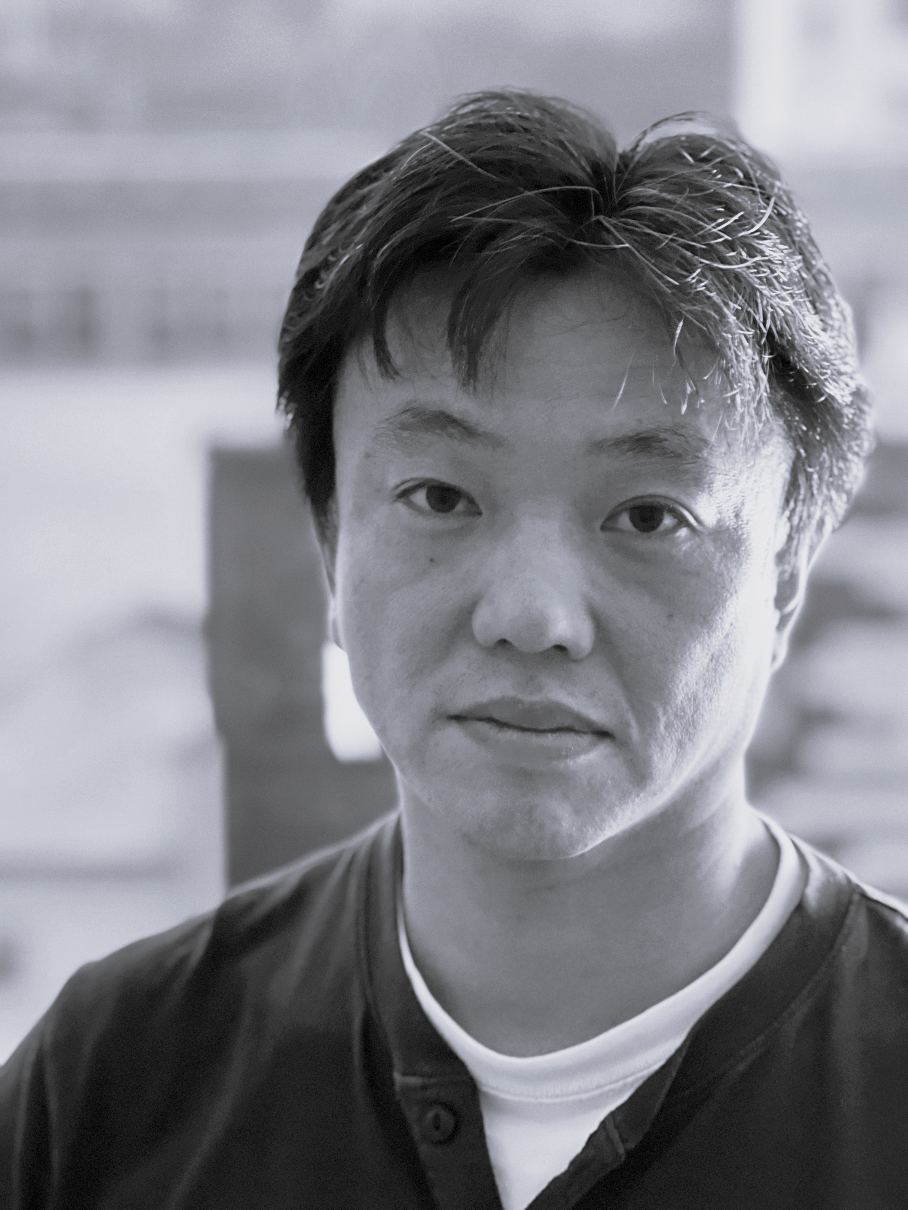}}]{Koichiro Niinuma}
is a Research Manager at Fujitsu Research of America, Pittsburgh, PA, USA. His research interests include computer vision, affective computing, biometrics, and human behavior understanding. He has over 50 U.S. and international patents, and his developed technologies, including patented technologies for fingerprint authentication and continuous authentication, have been used in many products. He received his M.S. degree from Kyoto University in 2003 and his Ph.D. from the University of Tsukuba in 2021.
\end{IEEEbiography}
\vspace{11pt}

\begin{IEEEbiography}[{\includegraphics[width=1in,height=1.25in,clip,keepaspectratio]{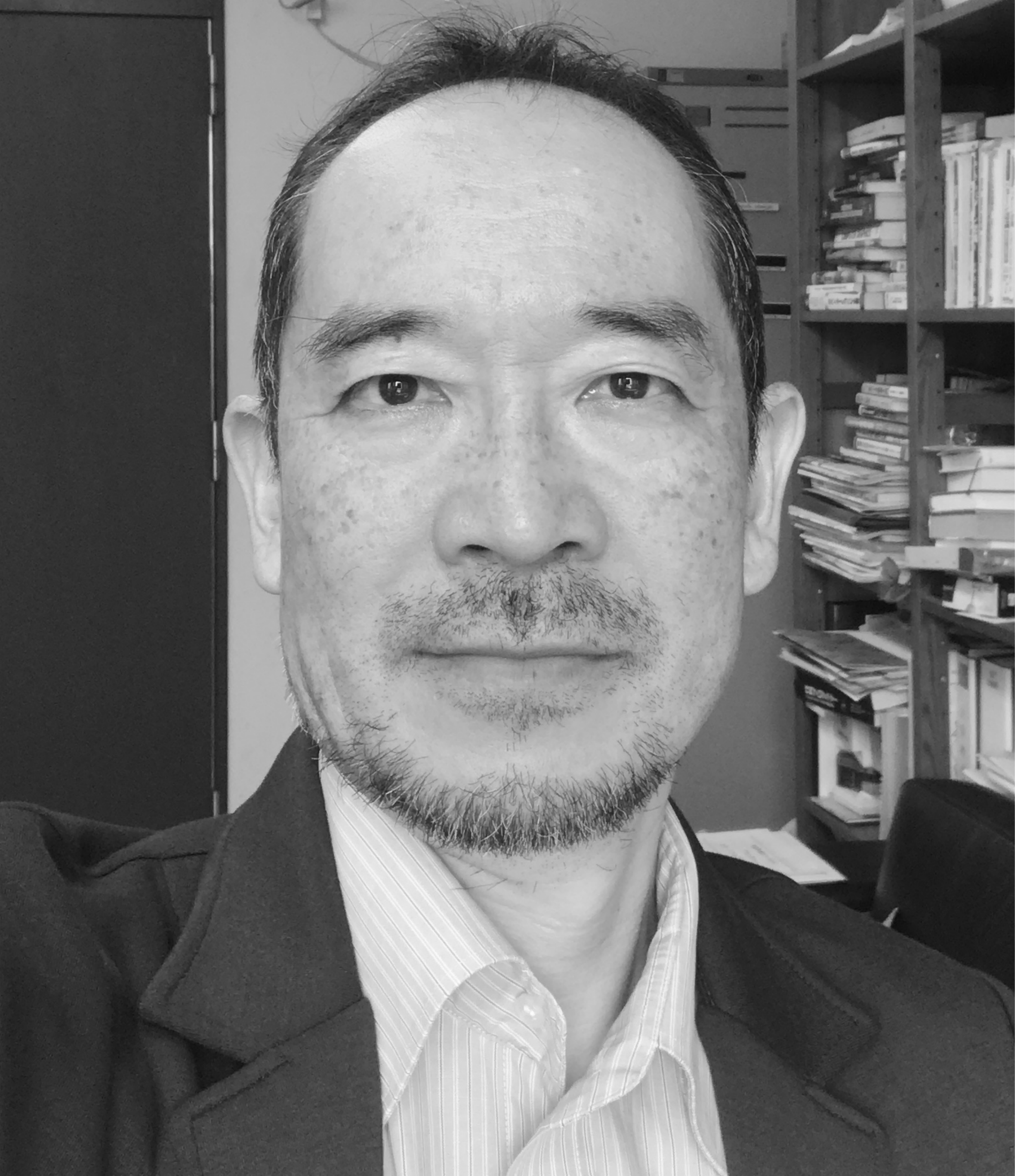}}]{Kazuhiro Fukui}
received his M.E. (Mechanical Engineering) from the Kyushu University in 1988. He joined Toshiba Corporate R\&D Center and served as a senior research scientist at Multimedia Laboratory. He received his PhD degree from the Tokyo Institute of Technology in 2003. He is currently a professor in the Faculty of Engineering, Information and Systems at the University of Tsukuba. His research interests include the theory of machine learning, computer vision, pattern recognition and their applications. He has served as a program committee member at many pattern recognition and computer vision conferences, including as an Area Chair of ICPR'12, 14, 16 and 18. He is a member of the SIAM.
\end{IEEEbiography}

\vfill

\end{document}